\documentclass[10pt, a4paper]{article}

\newcommand{\ange}[1]{\colorbox{teal}{\color{white}\textsf{A.}}
\textcolor{blue}{#1}}

\newcommand{\comments}[1]{\colorbox{teal}{\color{white}\textsf{A.}} \textcolor{green}{#1}}

\newcommand{\xlsr}{\texttt{XLS-R-En}}
\newcommand{\wtvt}{\texttt{Wav2vec 2.0}}
\newcommand{\whisper}{\texttt{Whisper}}

\usepackage{placeins}
\usepackage{enumerate} 
\usepackage{enumitem}
\usepackage{bbm}

\makeatletter
\renewcommand{\fnum@figure}{Fig. \thefigure}
\makeatother

\usepackage{amsmath}

\usepackage{amssymb}

\usepackage{mathabx}

\usepackage{titlesec}
\titleformat{\subsubsection}[runin]
  {\normalfont\normalsize\bfseries}{\thesubsubsection}{1em}{}
  
\titlespacing*{\subsubsection}{0pt}{0pt}{2em}

\usepackage[final]{lrec2026} 

\title{Responsible Benchmarking of Fairness for\\ Automatic Speech Recognition}

\name{Felix Herron\textsuperscript{(1,2)}, Ange Richard\textsuperscript{\ddag (2,3)},\\ {\bf \large François Portet\textsuperscript{\ddag, (2)}}, {\bf \large Alexandre Allauzen\textsuperscript{(1)}}, {\bf \large Solange Rossato\textsuperscript{\ddag, (2)}} \thanks{\ddag \quad contributed towards formulating framework of speaker group intersectionality and multivariate SG's.}} 

\address{MILES Team, LAMSADE, Université Paris Dauphine-PSL (1)\\ GETALP Team, LIG, Université Grenoble Alpes (2)\\ PACTE, Université Grenoble-Alpes (3) \\
felix.herron@univ-grenoble-alpes.fr\\}

\abstract{
    Many studies have shown automatic speech processing (ASR) systems have unequal performance across speaker groups (SG's). However, the manner in which such studies arrive at this conclusion is inconsistent. To pave the way for more reliable results in future studies, we lay out best practices for benchmarking ASR fairness based on literature from machine learning fairness, social sciences, and speech science. We then perform a case study on the Fair-speech benchmark, applying aforementioned best practices, and discuss how failing to do so can result in erroneous conclusions. On the whole, we advocate for as fine-grained an analysis as possible, taking into account as many variables as are available, in order to eschew dataset-level bias.
    \\ \newline  }

\begin{document}


\maketitleabstract

\section{Introduction}


In recent years, automatic speech recognition (ASR) software has grown increasingly performant \cite{asr_survey_modern}, which has led to a complementary increase in prevalence of ASR use among diverse populations \cite{tech_prevalence_older_adults,who_uses_smart_speakers,smart_speakers_old_adults}. It is therefore increasingly imperative to ensure that existing ASR systems perform equally regardless of the identity of the speaker. There is ample research demonstrating that certain speaker groups (SG's), such as children and non-native speakers, are treated worse than others by ASR systems. However, the methodology for identifying such SG bias in ASR systems is inconsistent across studies and sometimes marred by lack of precise analysis of the multifaceted identities of individual speakers.


Indeed, applying fairness benchmarks without sufficient oversight can lead researchers to stumble into erroneous conclusions which are incongruous with real world biases, a common blunder in fairness research \cite{fairness_traps}. This paper discusses the importance of defining SG-level fairness as intentionally and precisely as possible to avoid such blunders. We start by diagnosing our observed lack of consistency across ASR fairness benchmarking studies, and suggest this is due in part to a lack of clarity about how SG-level fairness should be defined and measured. We then list several best practices to avoid accidentally measuring bias stemming from fairness corpora rather than real-world bias. We formally define fairness in ASR, as motivated by broader fairness literature in machine learning (ML). Finally, we perform a case study on a common fairness corpus, Fair-speech, and apply many of the best practices to our analysis. We highlight some pitfalls that could entrap unaware users of the corpus. We finish by suggesting how future work, particularly in dataset creation, can help facilitate fairness benchmarking in ASR.




\section{Motivation}
\label{sec:motivation}

This study is motivated by a lack of consistency both within single and among studies examining fairness in SOTA ASR systems. We were puzzled to find that some studies report that women experience \textit{significantly worse} treatment \cite{big_thorough_bias_survey,solange_balanced,bing_youtube_dialect,hend_gender}, \textit{significantly better} treatment \cite{meta_fair,towards,dutch_quantifying,arabic_asr}, or \textit{both}, depending on subgroup studied \cite{gender_perf_gaps_multilingual_pebbles,region_bias_scots,bias_environment}. Likewise, while most studies find that non-native speakers are \textit{less well understood} by ASR systems \cite{native_non-native_accent,using_dat_vc_dutch,sonos}, some \textit{find the opposite} \cite{meta_fair}. Studies seem to all agree that \textit{children receive worse performance} by ASR systems; however, whether older adults are better understood than younger \textit{varies broadly by publication} \cite{aman2013speech,dutch_quantifying,bias_environment,sonos}.

It is possible that this effect is due in some part to the different ASR models being evaluated by each of these studies. Fixing a dataset and two SG's $g_1, g_2$ (e.g. native vs non-native speakers), some studies find that different ASR models perform better on $g_1$ while others on $g_2$ \cite{gender_perf_gaps_multilingual_pebbles}. However, most studies in the literature find that different ASR systems tend to be biased against the same SG's \cite{towards,dutch_quantifying,dutch_wav2vec_whisper}. If we assume that all ASR systems reflect the same SG-level biases, we would hope that all studies into ASR fairness would arrive at similar conclusions. That they don't is therefore likely due to methodological variance which allows researchers analyzing the same data to arrive at different conclusions.

\section{Best practices in reducing transmission of dataset bias}

It is important to remember that any conclusions reached by fairness studies are \textit{estimates} of real life bias, influenced by data on which the experiments were performed. With this in mind, researchers should try and limit dataset-level bias as much as possible so that their estimates are as close as possible to a real world simulation. In this section, we will highlight several key notions to attenuate filtration of biases due to dataset construction into fairness results. For each, we cite examples from the literature. It is important to note that these best practices remain general - we see them as akin to tools in a belt, whereby the user still must know how to use each in the manner most beneficial to their use case. The subsequent section describes a case study on how these tools can be put to use - however, each setting is different.

\subsection{Ensure equal distribution of recording quality}

Background noise (and recording quality in general) have been shown to impact ASR performance \cite{noise_means_bad_asr}. It is possible that some SG's will have different levels of background noise, thus potentially biasing results of a fairness study. Some benchmarks circumvent this by recording all of their inputs in the same conditions \cite{sonos,meta_fair}, though benchmarks compiled from diverse sources cannot \cite{artie}. That said, it can be useful to have variably noisy recordings in the dataset, as real life ASR often occurs in noisy environments using old recording equipment. Furthermore, some SG's are more likely experience such potential impediments to ASR transcription, such as the "low" socio-economic status SG in the Fair-speech dataset \cite{meta_fair}.

\subsection{Verify text complexity}

If different SG's use different vocabulary/grammar (i.e. text) in the real world, it is important that these attributes be taken into account during bias testing. A corpus where all speakers speak comparable texts cannot be considered to faithfully estimate bias if this is not the case in the real world; likewise, if a certain SG in a corpus is comprised of recordings of complex texts, while another SG speaks easy texts, this likewise has the potential to engender unfaithful bias estimations. On the other hand, if one is trying to capture \textit{merely} the bias due to acoustic features of a speaker's voice, then controlling for text (complexity) is beneficial. Once again, this is a decision that researchers must consider intentionally in the context of their individual study.

For example, \citet{black_white_princeville} proposes calculating each text's perplexity to assess whether each SG speaks similarly complex sentences. They also measure the "dialect density" of text to determine the extent to which it contains grammar typical of African American Vernacular English.

\subsection{Understand intra-SG speaker diversity}

When we measure ASR performance w.r.t. a SG, it is tempting to treat SG labels as precise, immutable defining characteristics of speakers. However, speakers within a given SG can be diverse. It is therefore essential to understand how each SG defined, what it means to belong to multiple SG's at once, and SG's are balanced throughout the dataset. 

\subsubsection{Intersectionality} Traditionally, fairness in ML has focused on comparing between outcomes for single groups from the same demographic variables (DV's), such as between different races or sexes \cite{ibm_single_axis,fairness_survey_classic}. However, recently the application of \textbf{intersectionality} in ML fairness research has gained traction as a technique to more precisely gauge bias \cite{fouldsIntersectionalDefinitionFairness2019,intersectionality_ml_survey}. These studies argue that measuring fairness w.r.t. a single DV in isolation is insufficient; to best understand fairness, one must look at as fine-grained SG's as possible, comprised of as many DV's as possible. 

Intersectionality has its roots in the social sciences where \citet{intersectionality_first} defines it as discrimination faced by members of multiple marginalized classes at the \textit{intersection} of several groups, for example Black. \citet{intersectionality_ml_survey} emphasizes that treating heterogeneous groups as a monolith (e.g. all people from Pacific islands) can hide unfair treatment experienced by subgroups thereof (e.g. specific islands). \citet{fouldsIntersectionalDefinitionFairness2019} emphasizes the importance of prioritizing protected classes which are underrepresented in fairness benchmarks, as their discrimination can be more easily ignored than a large underprivileged group.

Several existing benchmarks for fairness in ASR already consider the intersectionality of SG's (without necessarily using that exact term) for both aforementioned motivations. For example, \citet{dutch_quantifying} and \citet{towards} examine the WER gap between Dutch spoken in Flanders vs in the Netherlands, intersecting with regional dialects, native-ness of speakers, age, and speech format (read vs. human-machine interaction). One finding is that the gender-based WER gap is least significant among children, while the age-based WER gap is most significant for women, as well as that the regional gap is strongest among children and teens, and weakest among older adults. This is a crucial insight that would be ignored if the authors had only compared between genders, ages, or regions.

\subsubsection{Conditional statistical parity}
\label{sec:parkinsons}

The metrics used to measure fairness in ASR correspond to \textbf{statistical parity} as introduced by \citet{fairness_definitions} in their landmark fairness taxonomy paper. A more rigorous version of this is \textbf{conditional statistical parity}, which requires that all secondary attributes about the setting be the same, such as background noise or text complexity. Furthermore, we can condition on/take into account secondary DV's in our calculations. This is important both in order to uncover intersectional biases, as well as to equilibrate potential unbalance in other DV's. Failure to do this might end up spuriously measuring the random side-effects of subgroup imbalance captured during dataset construction. 

As a toy example, let us imagine we are measuring the fairness of performance of an ASR system on men vs women, on some benchmark $B$. By chance, 1 in 20 men in $B$ suffer from Parkinson's disease (diminishing their ASR comprehensibility \cite{parkinsons_asr}), but only 1 in 100 women in $B$ suffer from Parkinson's. If we are either unaware of this, or ignore it, then we might conclude that men experience worse ASR performance than women; however, our observation would at least in part be due to the confounding influence of a higher prevalence of Parkinson's in the male population, rather than due to their masculinity.

\citet{black_white_princeville} responsibly attempt to avoid contamination of their race DV by either age and gender by retaining the same proportion of both gender and age groups for both White and Black speakers. However, they don't consider the intersection of age and gender, an oversight of intersectionality. They then delve into the geography of both racial groups where, crucially, they find that race is not sufficient to explain the WER gap; Black Americans from Rochester had comparable WER to White Americans.

\citet{sonos} control for the confounding effects of other DV's, noting that this causes some univariate effects w.r.t. age and gender to vanish. Likewise, \citet{region_bias_scots} finds a greater difference in by-gender performance in certain dialect groups.

On the other hand, \citet{meta_fair} finds that men have twice as high a WER as women, which they explain by citing previous work showing men tend to have worse ASR performance. However, while they note that the men in their dataset are far more likely to be African-American than their women, they don't perform an intersectional analysis to interrogate what is likely at least partially responsible for that effect.

Another example of failure to take SG diversity into account is the "Asian" dialect category in the Sonos dataset \cite{sonos}, which is comprised of speakers from Southern as well as Eastern Asian countries. The authors acknowledge the extraordinary diversity of this category and the incumbent challenges this causes in interpreting results based on it.

\subsubsection{Beware (un)known confounding factors}

In the previous examples, researchers were able (or failed) to avoid jumping to conclusions more reflective of biases in their dataset than biases in their ASR systems. Or they were able to uncover intersectional bias specific to multidimensional SG's. However, these effects can only be explicitly controlled when potentially confounding metadata are available in the dataset. For example, the Sonos dataset has ethnicity tags for only a small portion of its speakers, where they show it to have a statistically significant relationship with ASR error rate \cite{sonos}. However, they cannot control for equal ethnicity distribution over the rest of the dataset, and therefore cannot control this bias.

Indeed, there may be many other confounding variables related to speaker identity which are not included in the dataset. It is by definition impossible to directly control for these; however, authors could estimate the extent to which their datasets are free from such effects by using phonetic priors. For example, studies have shown that human's voices don't change very much during middle age \cite{old_voice_change}. The reliability of fairness experiments can therefore be benchmarked by performance variance across middle-aged age groups: if there is significant performance difference between any two middle-aged SG's, that is an indication of methodological error, likely due to lack of balancing \cite{sonos,meta_fair}. 




\subsection{Define SG-level performance based on speaker-level performance}
\label{sec:speaker_not_utterance}

When measuring SG-level bias, it is imperative to calculate error for each SG as a function of error for each speaker adhering to said SG. This is based on two observations: first, utterances from the same speaker are not independent, and thus we cannot perform a statistical test that assumes independence of samples. Second, this avoids bias due to imbalance in representation for each speaker. For example, if speaking time is not equally distributed across speakers in SG (e.g. $D'_{SG}$ contains many more utterances/words for some speaker $S_1$ than another speaker $S_2$), then calculating SG-level error as a function utterances will engender bias towards Speaker $S_1$, and will not be a faithful representation of the SG overall. 

Not all studies adhere to this principle, however. \citet{sonos} explicitly argues for the simplicity of measuring fairness based on individual utterances. Furthermore, \citet{dutch_quantifying} and \citet{towards} base some of their conclusions on a small numbers of speakers (see i.e. Table 1 in \citet{towards}). They claim statistical significance, likely based on the number of overall samples or hours of recorded speech, rather than the diversity of speakers per SG.

\subsubsection{The challenge of speaker paucity}
\label{sec:cohen}

If we define SG-level performance as a function of individual speakers, the statistical significance of our results will depend on the number of speakers in each SG. This leads us to a set of contradictory incentives: the more precisely we define SG's as the intersections of multiple DV's (as encouraged in the previous section), the more precise are our conclusions into SG-level fairness. However, the more precisely we define SG's as the intersections of multiple DV's, the fewer speakers will be included in each class, thus reducing the significance of tests we perform on them. This is logical: if a SG contains too few speakers, we risk measuring bias due to the unique nature of those several individuals, rather than due to the SG they belong to.

We can derive the number $n$ of speakers per SG necessary for statistical (with confidence $\alpha$ and power $\beta$) significance in, for example, a one-sided two-sample Z-test (e.g. comparing the mean error of two SG's given fixed population variance) with test-statistic $Z$:

\begin{align}
    &Z := \frac{\hat{\delta}}{\sigma\sqrt{2/n}} \nonumber\\
    &Z > z_{\alpha} + z_{\beta} \implies \frac{\hat{\delta}^2}{\sigma^2*(2/n)} > (z_{\alpha} + z_{\beta})^2 \nonumber\\
    & \implies n > 2*\frac{(z_{\alpha}+z_{\beta})^2\cdot \sigma^2}{\hat{\delta}^2} \label{eq:cohen}
\end{align}

\noindent where $\hat{\delta}$ is the difference in estimated WER for two SG's, $\sigma$ the variance between speakers, and $z_{\alpha, \beta}$ the quantiles defined by the significance and power respectively. For example, given typical values of $95\%$ confidence with $0.8$ power, taking $\hat{\delta} = 0.1$ and $\sigma = 0.15$ (reasonable estimates based on our analyses in Section \ref{sec:analysis}), we would need $n \approx 35$ speakers per SG. This could be a serious hindrance for corpora with few speakers, and/or multivariate SG's defined by many DV's.

This bound cannot be shrunk simply by increasing the number of utterances per speaker. $\hat{\sigma} := \sigma + \epsilon$, where $\epsilon > 0$ varies inversely to the number of words for each speaker - the more words available per speaker, the smaller $\epsilon$ and thus the less noisy $\hat{\sigma}$, which results in smaller $n$. However, $\hat{\sigma}$ can never be lower than $\sigma$, which means that an increase in the number of words per speaker has a limited effect on improving our measured error bounds.

\subsection{(Sometimes,) aggregate SG's}
\label{sec:aggregation}

Just because metadata are available in a corpus doesn't mean they are useful in fairness analysis! 


\subsubsection{Too few speakers per SG} In this case, we will be unable to draw statistically significant conclusions based on performance over this SG. Therefore, it could make sense to create an "other" category which groups semantically unusual SG's together. The upside of such aggregation is that it allows "other" to be represented by sufficient speakers so as to be statistically significant, whereas as initially constituted, those SG's would be statistically meaningless. The downside is that this marginalizes the individual identities of those underrepresented SG's, which \citet{intersectionality_ml_survey} warns against. However, we are limited by the data available, and there is interpretable value in creating a class to compare with the mode SG's.

\subsubsection{Superfluous level of precision in metadata} We stand to gain nothing by measuring fairness w.r.t subgroups defined by attributes likely independent of ASR fairness, for example different middle-age subgroups \cite{old_voice_change,children_development_age_2}. Instead, this can lead to two harms: 1) we risk accidentally creating SG's which are unbalanced w.r.t. other underlying SG's, thereby potentially corrupting our analysis. 2) it reduces the number of speakers in each multivariate SG, thereby lowering the statistical significance of results. Thus, we might consider aggregating all middle-aged speakers into the same SG.

\subsection{Outlier speaker removal}

A final source of bias potentially contaminating our SG-level fairness results are outlier speakers. If we want to want to measure the general behavior of a certain SG, which potentially contains few speakers (due to dataset constraints), and one of the speakers is understood much worse than the rest, that is potentially due to some individual speaker-level characteristics which are not relevant to our analysis. It can therefore make sense to exclude extrema values from SG-level mean calculation, for example. This is less of a problem for datasets with very high numbers of speakers; however, that is unfortunately rarely the case in practice.

\section{Quantifying fairness in ASR}
\label{sec:methodology}

We describe the two metrics most often used in the literature to quantify bias in ASR systems.

\subsection{Relative SG-level error/WER gap}

Most studies into SG-level ASR fairness measure the relative error rate for each SG. For a dataset $D$, they calculate the word error rate (WER) for each utterance, a measurement of the number of substitutions, deletions, and insertions necessary to correct the automatic transcription by an ASR model $M$ over some $D' \subseteq D$. Some studies then calculate the average WER for each SG as the average WER for all utterances $D'_{SG} \subseteq D'$:

\begin{align}
    &\text{WER avg.}^*(D'_{SG}) := M^*(D'_{SG}) := \label{eq:avg_perf_bad}\\ &\frac{1}{|\{u \in D'_{SG}\}|} \sum\limits_{u \in D'_{SG}} \text{WER}(u; M) \nonumber
\end{align}

\noindent However, as mentioned in Section \ref{sec:speaker_not_utterance}, Metric \ref{eq:avg_perf_bad} falls immediately into a hazard of imprecise measurement by failing to first average by speaker. A more prudent approach is:

\begin{align}
    &\text{WER avg.}(D'_{SG}) := M(D'_{SG}) := \label{eq:avg_perf}\\ &\frac{1}{|\{S \in SG\}|} \sum\limits_{S \in SG} \frac{1}{|\{u \in D'_{S}\}|} \sum\limits_{u \in D'_{S}} \text{WER}(u; M) \nonumber
\end{align}

\noindent Then, one can measure bias against a particular SG $SG_i$ in terms of the relative performance between $SG_i$ and some the rest of the subset $D' \subseteq D$ (typically, studies use $D'=D$) \cite{meta_fair,towards}:

\begin{align}
    \text{Err}_{rel}(SG; D', M) := 100 \times \frac{M(D'_{SG}) - M(D')}{M(D')} \label{eq:rel_perf}
\end{align}

Some studies also calculate the unfairness w.r.t. a DV as the difference between the best and worst relative error for two constituent SG's, often denoted the WER gap \cite{hend_gender,gender_perf_gaps_multilingual_pebbles,fairspeech_bad_methodology_asr}:

\begin{align}
    \text{Unfairness} &:= \text{WER gap}(DV; D', M) := \label{eq:unfairness}\\ &\max_{SG_i \in DV}\{\text{Err}_{rel}(SG_i; D', M)\}\nonumber\\ - &\min_{SG_j \in DV}\{\text{Err}_{rel}(SG_j; D', M)\} \nonumber 
\end{align}

\noindent Then, one can perform a statistical test, such as a 1-sample (or 2-sample, if comparing between pairs of SG's rather than with respect to the dataset on average) t-test, to determine whether the relative error and unfairness are statistically significant for SG's and DV's respectively. An ASR model is deemed fair w.r.t. a SG if it delivers statistically negligible relative error, and fair w.r.t. a  DV if it delivers a statistically negligible WER gap.

\subsubsection{Isolated effect of single DV's}
\label{seq:isolated}

As mentioned in Section \ref{sec:motivation}, we recognize that individual SG's potentially contain heterogeneous subgroups. Thus we seek to isolate the effect of individual DV's on ASR performance. For any $DV^i \in [DV^1,..,DV^k]$ (e.g. gender), one of many DV's included in metadata, we define a subset $D'_{cond, i} \subseteq D$ by fixing a $SG_{DV_j}$ for every other $DV_j \in [DV^1,..,DV^k] \setminus \{DV_i\}$ (e.g. only children, only non-native speakers, only African Americans, etc.):

\begin{align}
    D'_{cond, i} = \bigcap\limits_{j \in [1..k]\setminus \{i\}} D_{SG_{DV^j}} \label{eq:d_cond}
\end{align}

\noindent and calculate $\text{Err}_{rel}(SG; D'_{cond, i}, M)$ for all $SG \in DV^i$ as well as $\text{WER gap}(DV^j; D'_{cond,i}, M)$. We can then aggregate mean performances for each permutation of other the other $DV^j$ and perform a statistical test, such as 1-sample t-test, to determine, w.r.t. a $SG \in DV^i$, whether the means of the relative error rates were statistically significantly different from zero, or w.r.t. $DV^i$ overall, whether the unfairness levels were statistically significantly greater than zero.

\subsubsection{Worst treated multivariate SG's}

We define \textit{multivariate SG's} as the intersection of every DV available in the dataset. For example, if the dataset is annotated with gender, age, and native language, an multivariate SG might be "female children who speak native English". For each multivariate SG, we calculate the relative error w.r.t. the dataset overall (Eq. \ref{eq:rel_perf}); then observe which multivariate SG's, if any, have the lowest/highest relative performance. This will uncover SG's whose marginalization is compounded by their intersectionality, as proposed by \citet{intersectionality_ml_survey}.

\section{Case study on Fair-speech}
\label{sec:analysis}

Unfortunately, many of the most prominent corpora for evaluating ASR systems do not permit bias evaluation due to a lack of sufficient recorded demographic metadata \cite{fleurs,librispeech,callhome_corpus} or unreliable labeling thereof \cite{common_voice,globe}. However, there are several corpora specifically designed for bias/fairness evaluation of ASR systems whose multitudinous metadata categories permit finer-grained ASR evaluation. We proceed to analyze the Fair-speech corpus \cite{meta_fair} and discuss how to implement some of the best practices from the previous section.

We replicate each experiment using three different near-SOTA ASR models: Whisper-medium (\whisper{}), wav2vec2-large-960h-lv60 (\wtvt{}), and wav2vec2-large-xlsr-53-english (\xlsr{}). \whisper{} was trained end-to-end for ASR on 680k hours of YouTube transcripts \cite{whisper}; \wtvt{} was pretrained on 60k hours of LibriLight \cite{wav2vec2}; \xlsr{} was pretrained on a multilingual corpus comprising 53 languages \cite{w2v2-xlsr}. \wtvt{} was finetuned on 960h of LibriSpeech \cite{librispeech}; \xlsr{} finetuned on the English split of CommonVoice \cite{common_voice}. We can thus test whether our results are specific to one architecture/training set, or general across ASR systems.

\subsection{Dataset description}

\begin{figure}[h]
  \includegraphics[width=\linewidth]{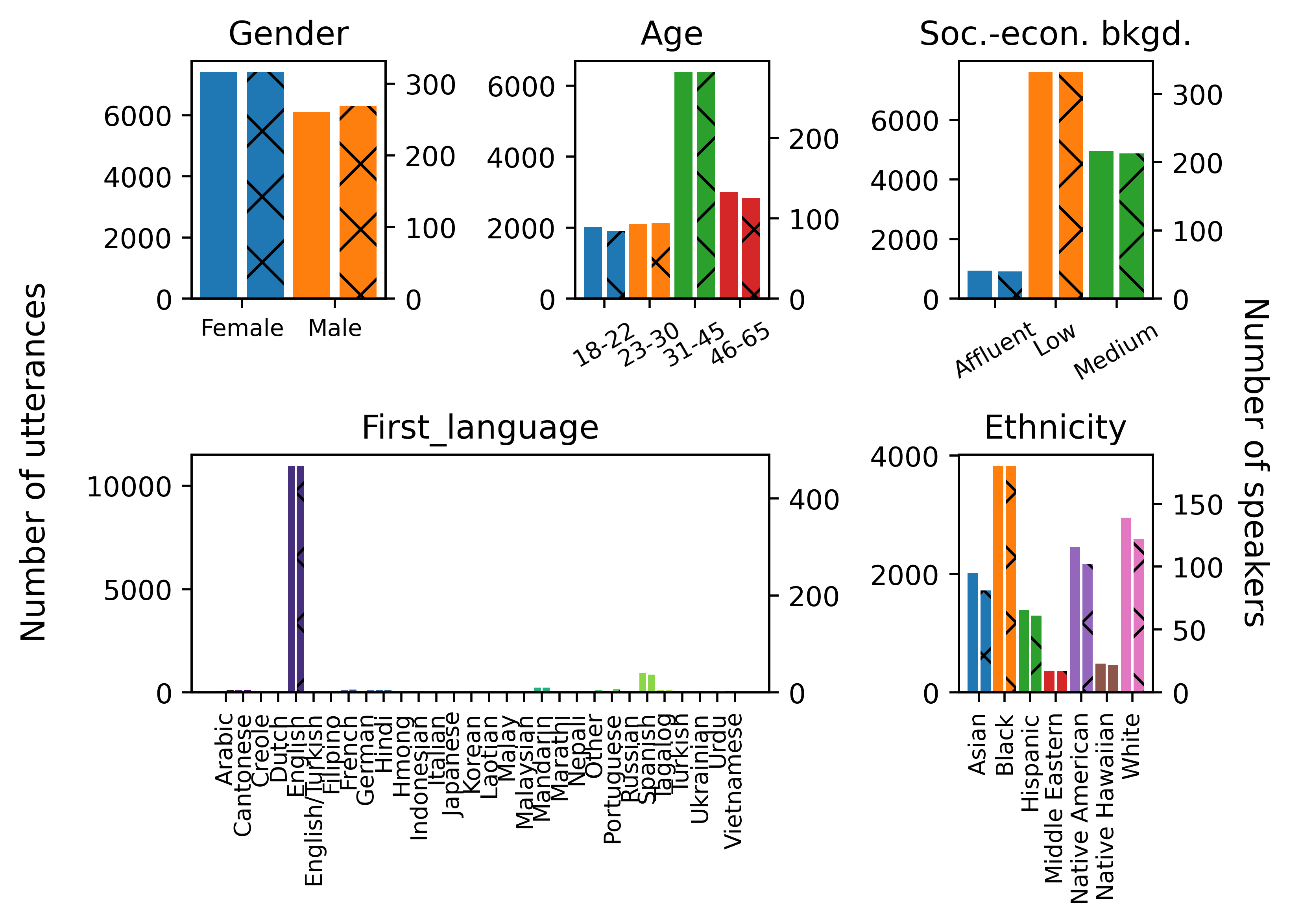}
  \caption{Overall distribution of demographic labels in the Fair-speech corpus. Note that most speakers are native English speakers, with a small (illegible) minority of Spanish native speakers.}
  \label{fig:meta_fair_demographics}
\end{figure}

The Fair-speech Dataset (Fair-speech) comprises 593 speakers over 56 hours \cite{meta_fair}. Fair-speech is comprised of recordings of paid speakers speaking (not reading) smart speaker commands. Speakers self-report metadata including: gender, age, ethnicity, first language, and socioeconomic background. See Fig. \ref{fig:meta_fair_demographics}.

\begin{figure}
  \includegraphics[width=\linewidth]{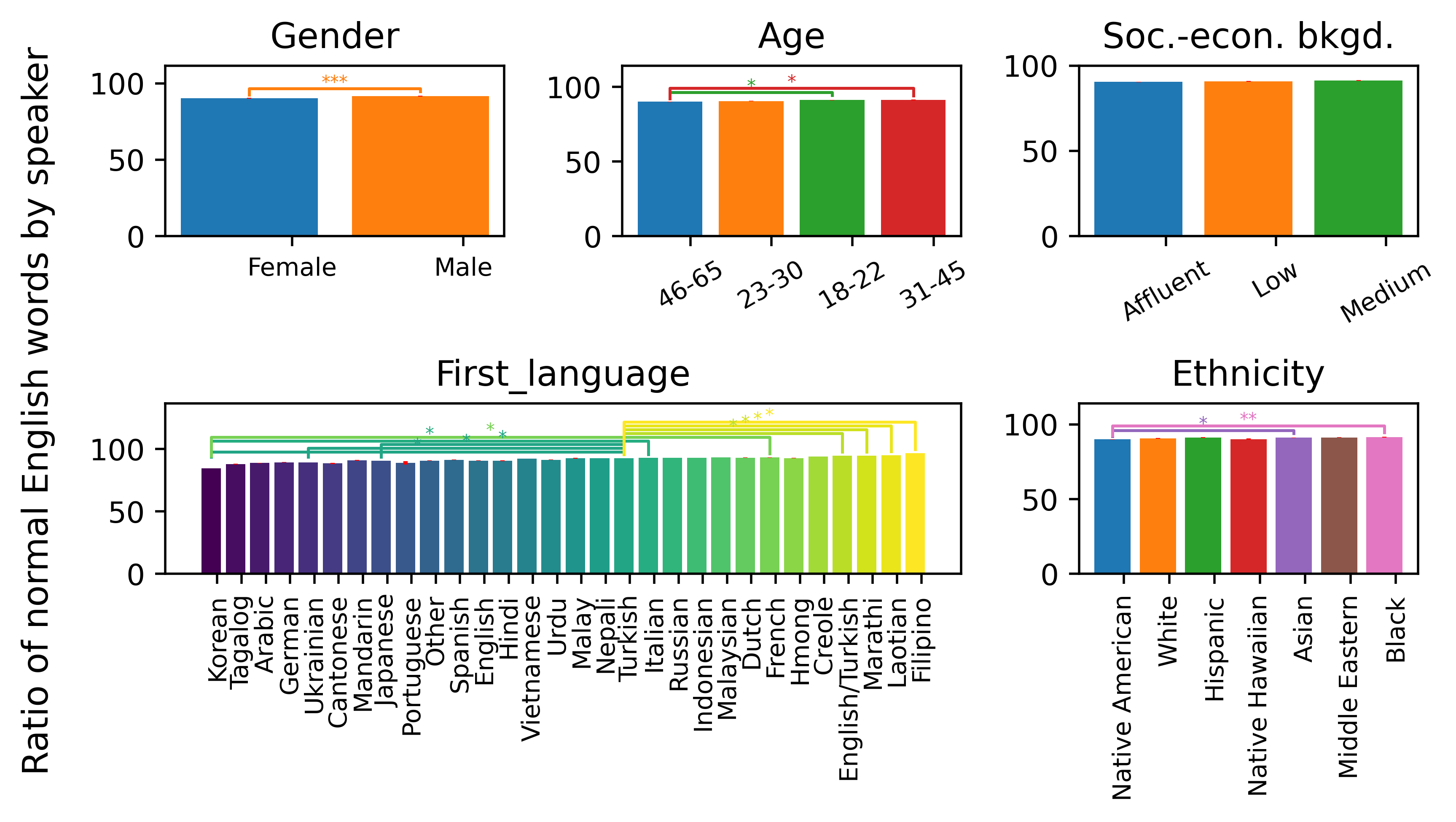}
  \caption{Ratio of non-English words per sentence, averaged by speaker.}
  \label{fig:meta_fair_sent_ratio}
\end{figure}

\begin{figure}
  \includegraphics[width=\linewidth]{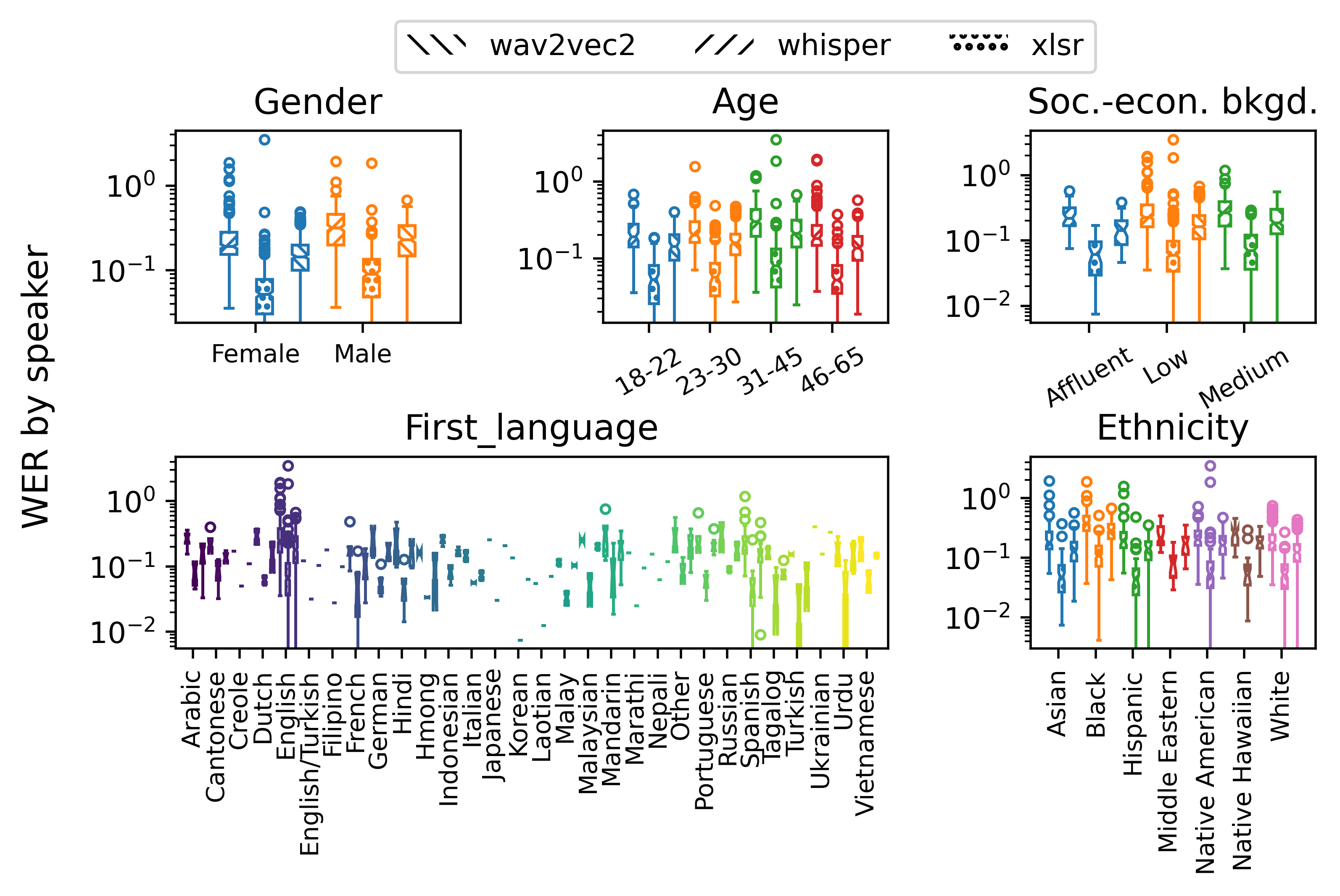}
  \caption{Variance of WER for each ASR model, averaged by speaker.}
  \label{fig:meta_fair_speaker_variance}
\end{figure}

\begin{figure}
  \includegraphics[width=\linewidth]{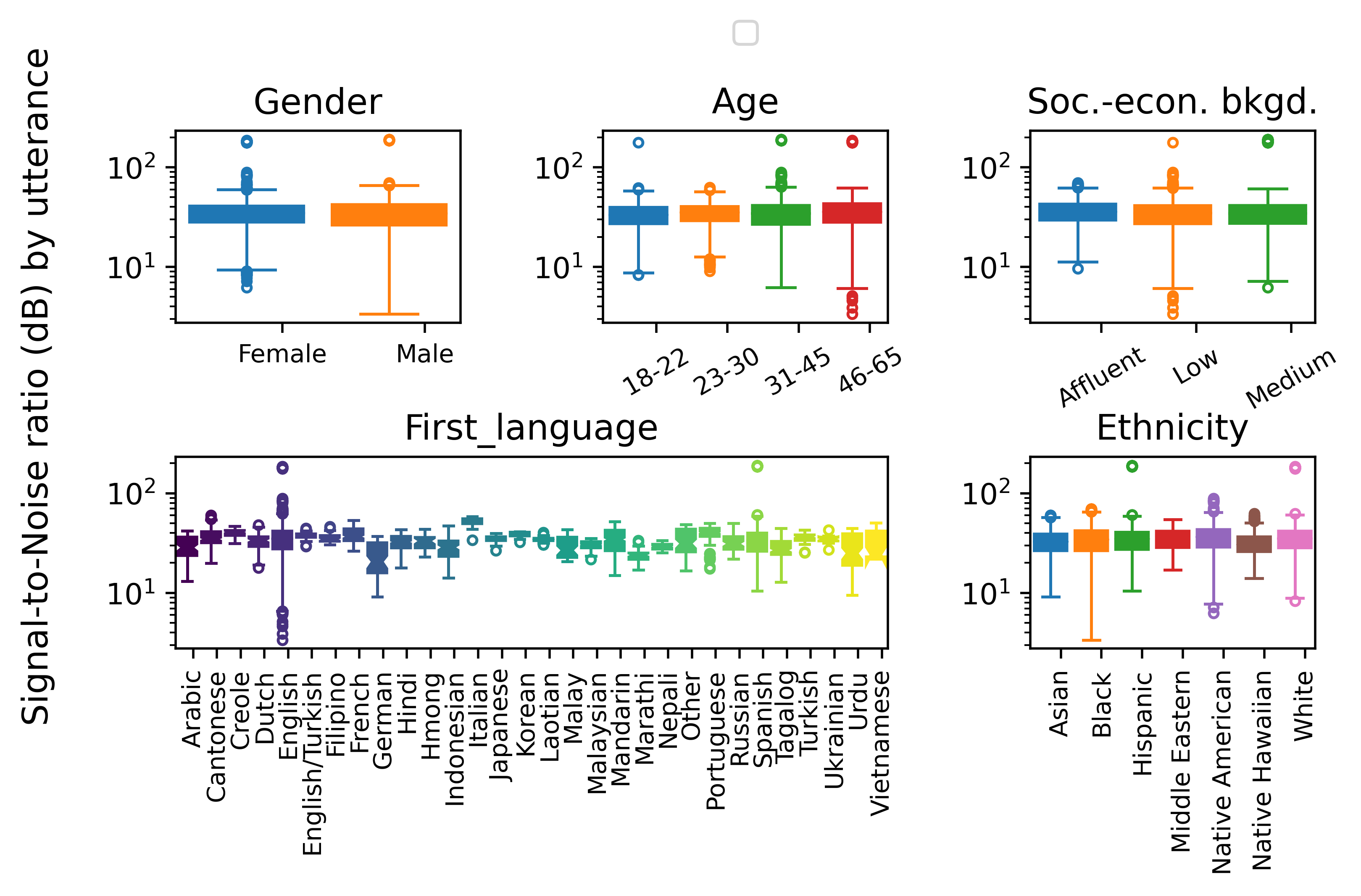}
  \caption{Signal-to-noise ratio for each recording.}
  \label{fig:meta_fair_snr}
\end{figure}



Fair-Speech represents many first languages; however, given a limited number of speakers per language, that might limit the statistical significance of results pertaining to speakers of sparsely represented languages. Furthermore, the number of age categories is likewise too precise - we likely stand to gain little by differentiating between different classes of middle-aged adults. Fair-speech lacks children and old adults, the two age-related SG's with phonetic motivation for divergent ASR error rates \cite{children_development_age_2,old_voice_change}.

\begin{figure}[h!]
  \includegraphics[width=\linewidth]{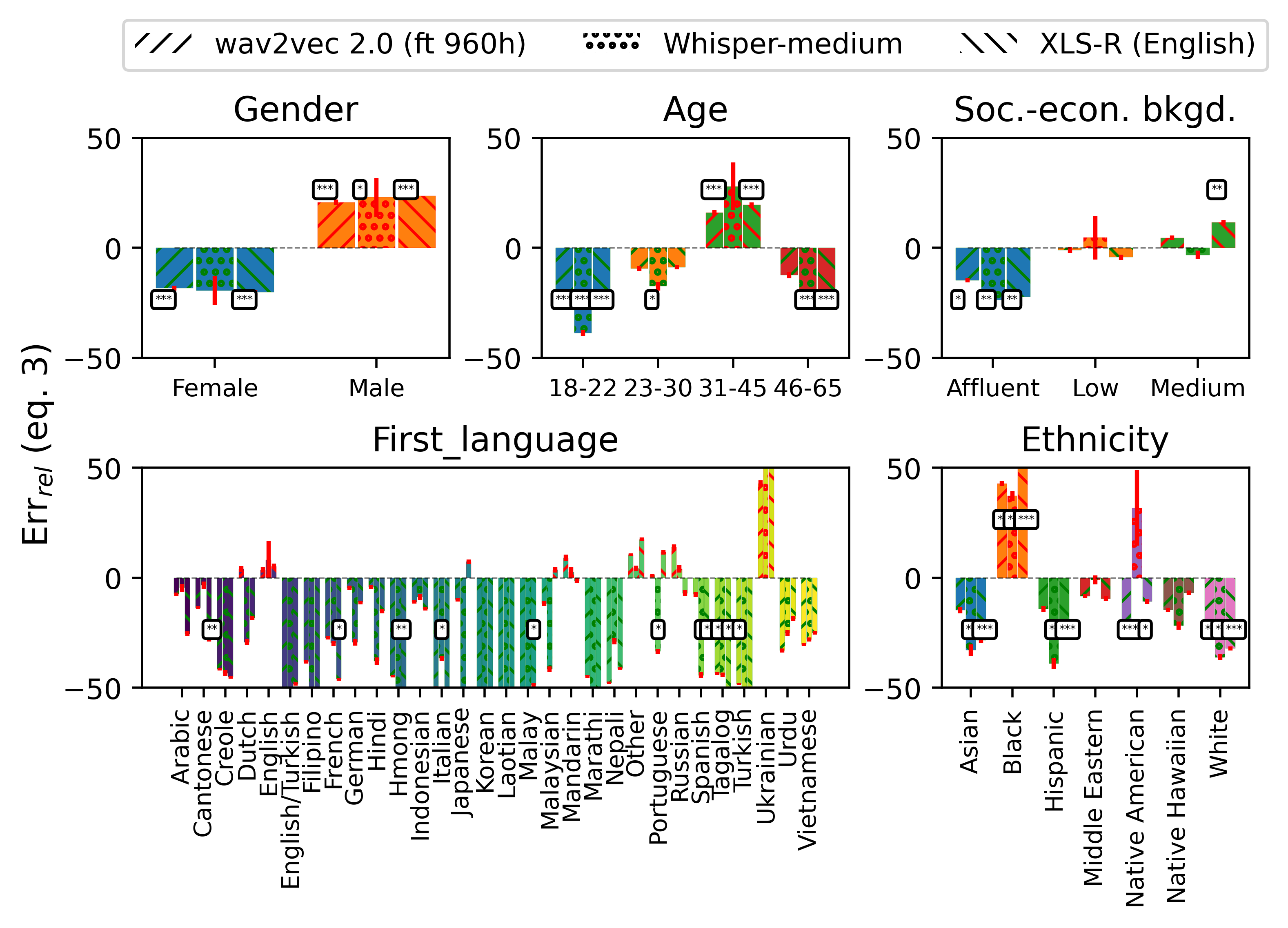}
  \caption{Fair-speech when measuring relative WER gap between over SG's belonging to a single DV at once. Values $<0$ indicate below average WER, i.e. above average performance. * denotes statistically significantly greater/less than 0 according to a 1-sample t-test - * implies $p < 0.05$, ** implies $p < 0.01$, *** implies $p < 0.001$.}
  \label{fig:meta_fair_original}
\end{figure}

\begin{figure}[h!]
  \includegraphics[width=\linewidth]{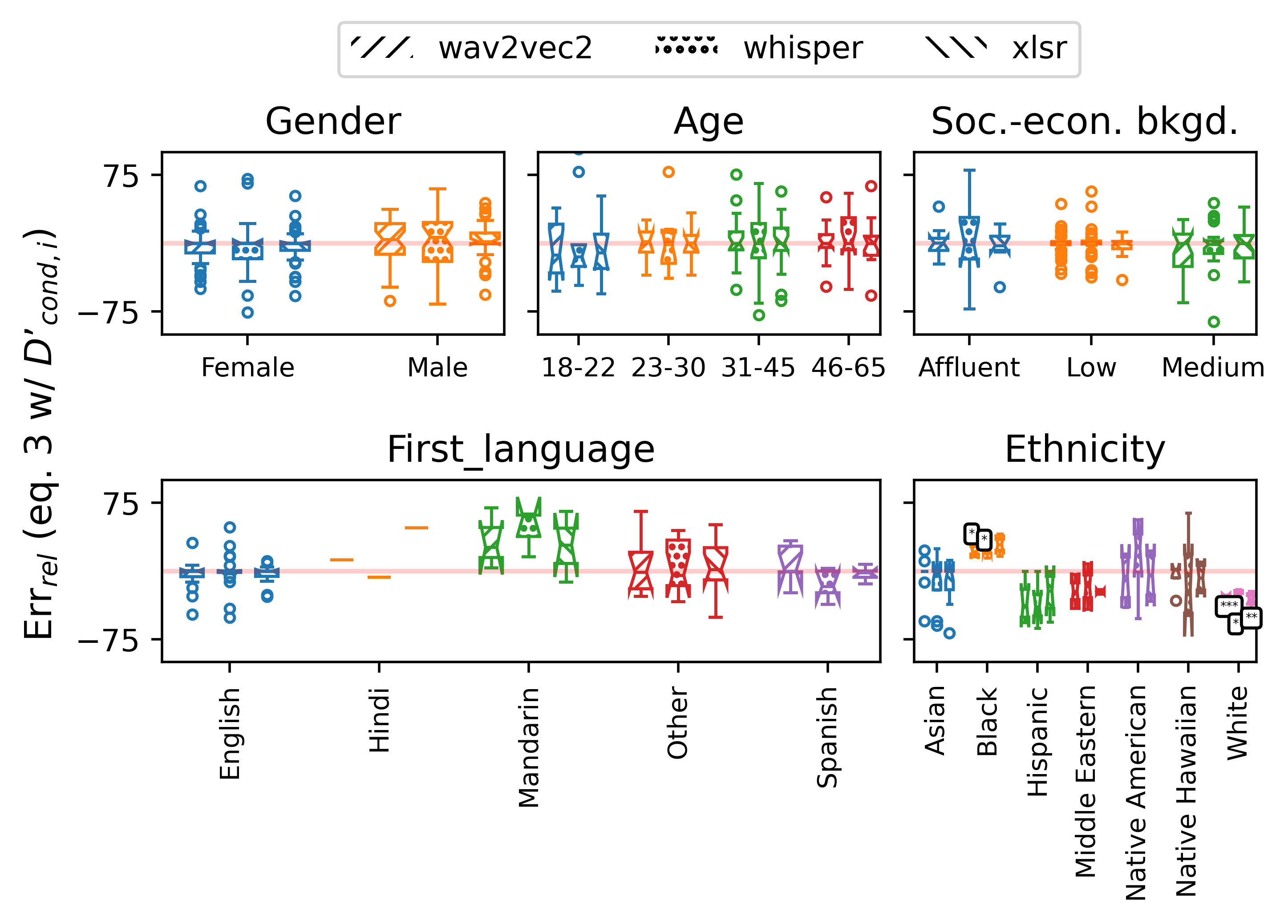}
  \caption{Fair-speech when measuring relative WER gap between intersectional SG's differing only on one specific DV. Each datapoint is a statistically significant difference between SG's differing by only one DV. Values $<0$ indicate below average WER (above average performance). * denotes statistically significantly greater/less than 0 in the aggregate according to a 1-sample t-test. * implies $p < 0.05$, ** implies $p < 0.01$, *** implies $p < 0.001$.}
  \label{fig:meta_fair_intersectional}
\end{figure}

\begin{figure}
  \includegraphics[width=\linewidth]{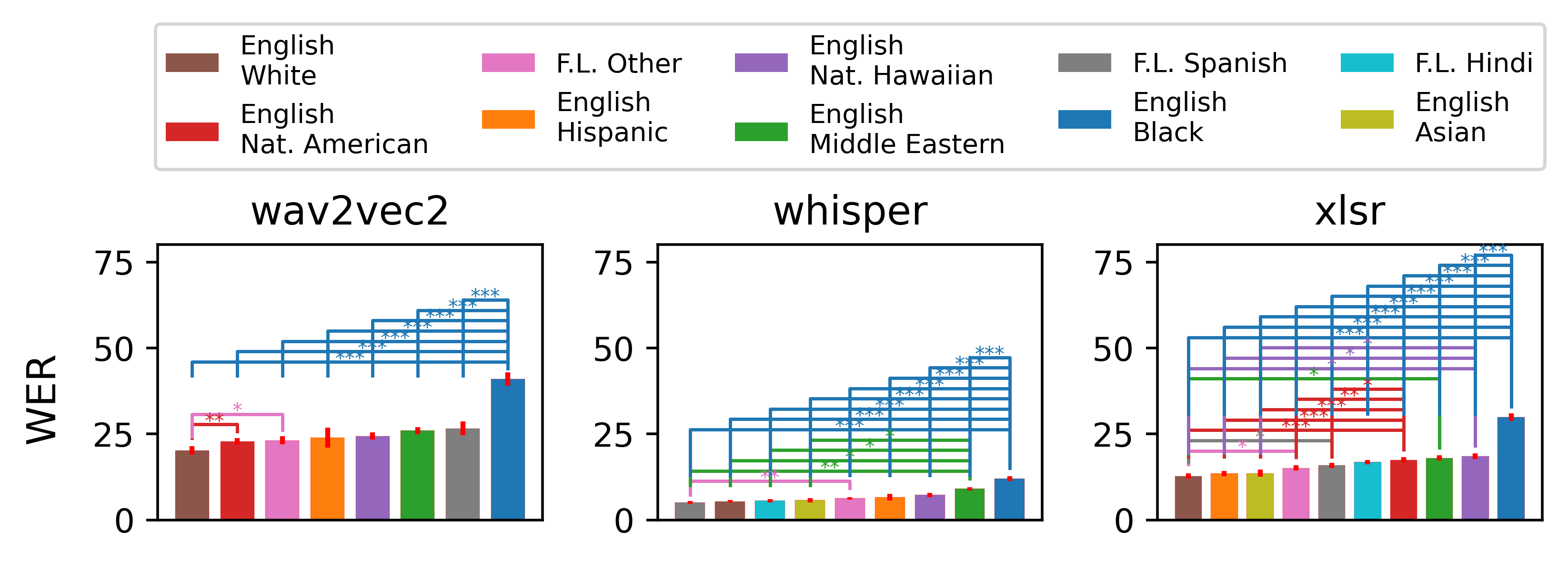}
  \caption{Overall WER for Fair-speech when conditioning on first language and ethnicity. * implies denotes significantly higher WER on a one-sided, two-sample two-sample t-test (* implies $p < 0.05$, ** implies $p < 0.01$, *** implies $p < 0.001$).}
  \label{fig:meta_fair_firstLang_ethn}
\end{figure}

\begin{figure*}
  \includegraphics[width=\linewidth]{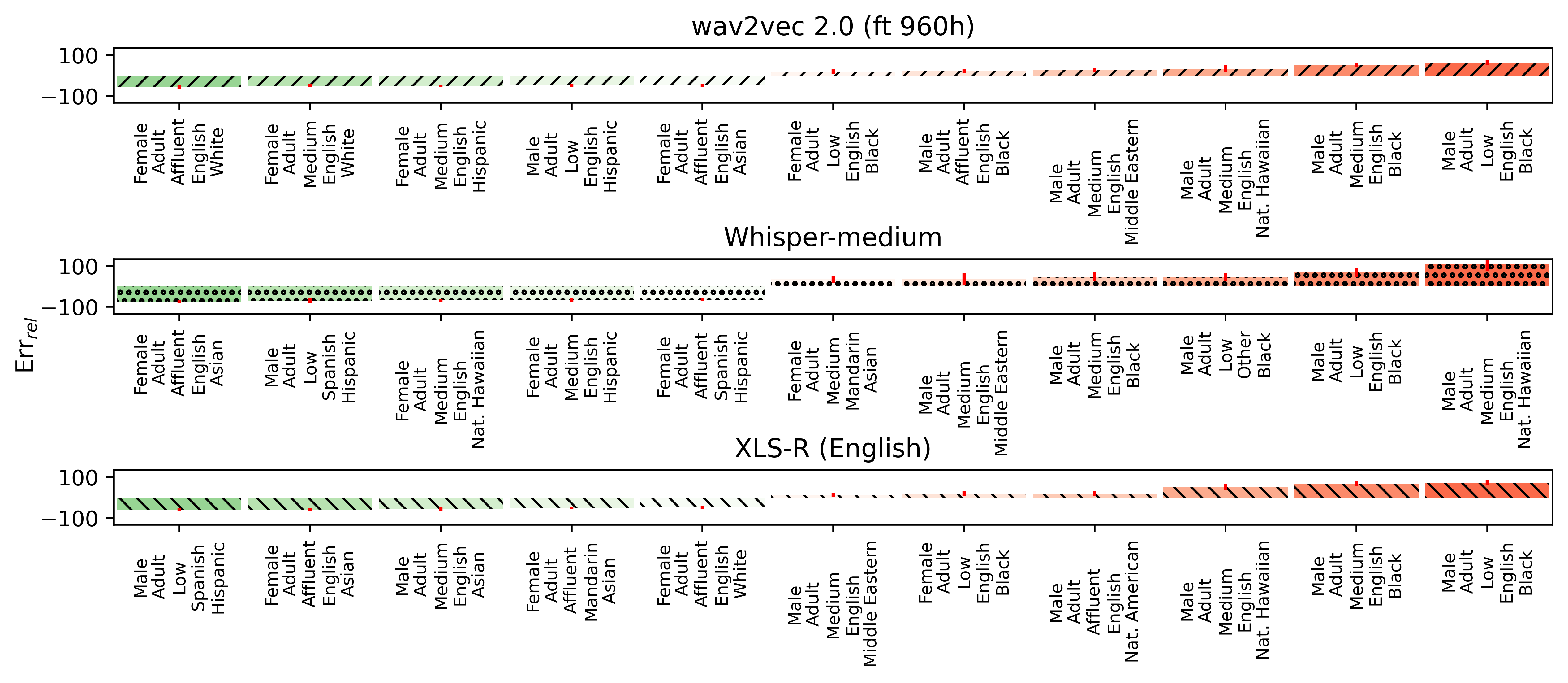}
  \caption{Relative error of intersectional SG's in Fair-speech with least, greatest WER (conditional on sufficient speakers w.r.t. Eq. \ref{eq:cohen}).}
  \label{fig:meta_fair_min_max}
\end{figure*}

\subsection{Filtering out outlier speakers and utterances}

First, we consider filtering out outlier speakers and utterances based on WER for each ASR model. Figure \ref{fig:meta_fair_speaker_variance} shows the variance of WER for each univariate SG and model. Note that some speakers have an average WER of over 1 - this is likely due to speaker-specific anomalies and not an accurate reflection of the ASR system overall. We proceed by filtering out all speakers (and utterances per speaker) with a z-score of $>3$ in each analysis that we conduct.

\subsection{Recording quality and text complexity}

Figure \ref{fig:meta_fair_sent_ratio} shows the average text complexity for every speaker, measured by number of words in the transcript that are not standard English (we use NLTK English dictionary \cite{nltk}). Overall, there is little variance, particularly among SG's with high representation; that said, the most disadvantaged SG's, as we will see in later analyses, are not necessarily those with the lowest ratio of standard vocabulary.

Figure \ref{fig:meta_fair_snr} shows the signal-to-noise ratio of each utterance, broken down by SG. Note that most recordings have are 10 dB (a reasonable threshold for ASR performance \cite{snr_asr_mobile}). For our experiments, we will remove all recordings at $<10 dB$.

\subsection{Calculating relative WER for each SG}
\label{sec:isolated}

We begin our case study by analyzing the results from Fair-speech. First, in Figure \ref{fig:meta_fair_original}, we present relative error rate as the dataset was constructed, without conditioning or manipulation. We measure statistically significant performance discrepancies w.r.t. each of the five DV's recorded in Fair-speech. Our results correspond to what was initially published in \cite{meta_fair}. Several odd results stand out, which raise some red flags about our experimental setup:

\begin{enumerate}
    \item 31-45 year-old's have higher WER than all other age groups. This is likely evidence of poor subgroup balancing, as there is no logical reason for different age groups of middle-aged adults to have variant performance.
    \item Men have vastly higher WER than women. \citet{meta_fair} attempt to explain the gender discrepancy by citing previous work showing men tend to have worse ASR performance; however, 100\% worse is much higher than peer studies \cite{sonos,hend_gender,gender_perf_gaps_multilingual_pebbles}.
    \item Most first languages have statistically insignificant relative WER. This is due to those SG's not being represented by enough speakers in Fair-speech.
    \item Native English speakers have negligibly higher worse-than-average WER, while several non-native speakers have statistically significantly better-than-average WER. This stands in contrast to most peer studies \cite{towards,dutch_wav2vec_whisper,sonos,native_non-native_accent}. This is potentially an artifact of disregarding intersectionality of multivariate SG's.
    
\end{enumerate}


\subsubsection{Analysis of multivariate SG's can help} Issue 1 is a warning that our "age" DV is probably improperly balanced w.r.t. other DV's. We reiterate that there is no good reason to analyze separate age groups of middle-aged adults; however, to investigate the hypothesis of imbalance in other DV's, we propose comparing only multivariate SG's where all DV's are the same except for age (see Section \ref{sec:isolated}). Figure \ref{fig:meta_fair_intersectional} shows that in doing so, the divergent performance w.r.t. age vanishes. That said, we note that the distribution for each class contains many outliers - this effect would likely diminish if we raised our threshold for number of speakers per multivariate SG. On the whole, this experiment supports our balancing hypothesis and supports the technique of multivariate comparison to avoid DV imbalance.

Issue 2 is similar to issue 1, and is similarly alleviated when considering only multivariate SG's differing only by gender (see Figure \ref{fig:meta_fair_intersectional}). As before, we note that the variance for both men and women is rather high - for some multivariate SG's, women have much higher WER than men, and vice versa. This  further motivates in-depth analysis of individual multivariate SG's to uncover potential intersectional SG's with compounded ASR error.

We note that the ethnicity DV is the only one to deliver statistically significantly different results - for Black and White speakers. Based on this experiment, we conclude that being Black is associated with inferior ASR performance across every permutation of other DV's, while the opposite is true for being White. This is a stronger conclusion than what we were able to draw from Fig. \ref{fig:meta_fair_original}.

\subsubsection{Conditioning on all DV's is no silver bullet}

Conditioning on all DV's allowed us to draw strong conclusions about Black and White speakers; furthermore, we can reasonably rule out gender, age, or socio-economic background having broad impacts on ASR performance. However, we cannot draw statistically significant conclusions about many first languages (issue 3) even after aggregating uncommon languages into "other", due to persistent lack of representation across multivariate SG's. Thus, equipped with our conclusions about insignificance of gender, age, and socio-economic background, we propose removing conditions on those three DV's, considering only difference in first language and ethnicity. Furthermore, for first languages which we retained after aggregation, we note that apart from English, there is one primary ethnic group that covers nearly all speakers of that language (all native Mandarin speakers are of "Asian" ethnicity, almost all native Spanish speakers are "Hispanic", etc). Thus, we move to condition on ethnicity \textit{only} for native English speakers.

Figure \ref{fig:meta_fair_firstLang_ethn} allows us to get a better sense of bias against multivariate SG's defined by first language and ethnicity. We find once again that Black native English speakers are less well understood than every other class in every model. One added insight is that not all native English speakers are treated the same - White speakers are the best native English speakers understood by every model, though not statistically significantly so for all ethnicities. This provides a clearer picture than the previous experiment which showed that White speakers were statistically always better understood than the mean. It also provides insight into issue 4, that non-native speakers are often better understood than native English speakers. When we condition on ethnicity, we find that it is only Black native English speakers that are worse understood, and that there is little statistical consensus regarding the relationship between being a native speaker and ASR error.

\subsubsection{Intersectional multivariate SG's with compounded ASR error}

Finally, we can analyze the intersectional SG's which have the least and greatest WER, which we show in Figure \ref{fig:meta_fair_min_max}. One surprising finding is that the worst performing group overall for \whisper{} is a Native Hawaiian group, which overall experienced much better treatment than Black speakers (Figure \ref{fig:meta_fair_firstLang_ethn}). One downside of this analysis, however, is that given the large difference in performance in extrema groups from the mean, fewer speakers are necessary to draw statistically significant conclusions based on SG (e.g. Eq. \ref{eq:cohen}). Thus, such fine-grained analysis has heightened risk of drawing erroneous conclusions.


\section{Discussion and outlook}

The primary takeaway from this work is an exhortation to future studies on fairness in ASR to be as fastidious as possible designing their experiments. We underscore the importance of an intimate understanding of the datasets on which one is evaluating before designing experiments, tailoring experiments to that data, and being transparent about limitations that can be drawn therefrom.

We also encourage authors to clearly delineate the questions which they seek to answer. On the one hand, we can estimate how individual DV's affect performance of ASR systems, like in RQ 2 - this is primarily useful for understanding ASR systems from a computational level and could help steer future work towards disaffected SG's in particular. On the other hand, we can estimate which SG's defined by the intersection of multiple DV's are treated the absolute best and worst by ASR systems. This gives us greater sociological insight into the ramifications of unfair ASR - if a speaker belongs to such a class, they risk acute discrimination.

Furthermore, we encourage humility on behalf of researchers into ASR fairness in the face of statistical uncertainty. As we describe in this study, current fairness benchmarks suffer from lack of speaker diversity. Using overly broad SG's reduces the narrative power of their analysis, while using overly precise SG's silos speakers into groups without enough speakers, rendering conclusions statistically insignificant. With these two goalposts in mind, given the constraints of current ASR fairness benchmarks, \textbf{the conclusions we can draw from this type analysis remains limited}. Indeed reporting a result as statistically insignificant or sociologically broad isn't a problem (as long as this is duly noted); rather, it is a reflection of the reality of limitations of current corpora. This leads to the obvious recommendation to collect more data with high-quality annotations. However, this is both expensive and ethically fraught - \citet{artie} explicitly avoids children, while some countries forbid the analysis of ethnic minorities \cite{french_data_collection_law}.


It is important to note that the DV's included in the three benchmarks which we studied are themselves sources of potential bias. Had the benchmark designers decided to record different metadata, our results would reflect that, both in our ability to observe both the SG's which trigger unfairness, as well as the intersectional SG's which maximize unfairness. Future work in defining ASR fairness datasets should be in consultation with sociologists and phoneticians to determine which DV's, and SG's therein, are a) important in the larger context of social discrimination and b) likely to contribute to disparate ASR performance. 

We encourage future study into linguistic or phonological mechanisms which are the actually underlying causal drivers of SG-level unfairness, beyond SG labels. Future work might eschew some of the pitfalls of relying on unbalanced and heterogeneous SG's advertised in this study by focusing on more rigorously defined measurements, such as dialect density measure \cite{black_white_princeville} or speed of speech \cite{mengDontSpeakToo2022}. Alternatively, unsupervised feature discovery has also been shown to uncover proxies for disadvantaged SG's in ASR without having to explicitly label them \cite{amazon_kmeans,laura_kmeans}. In other areas of ML, where fairness depends on sensitive attributes being evenly distributed in decision functions, there is work in automatically selecting attributes to include in fairness analysis \cite{sensitive_attributes_automatic_stats}. This approach takes SG intersectionality into account by testing which combinations of attributes are related to an outcome variable. 


One potential avenue for more precise fairness analysis is in conditional synthetic voice generation, such as \cite{ancogen_generation}. This allows for two utterances from two different speakers to be \textit{exactly} the same, apart from some parameters that are meant to mimic particularly DV's. For example, \citet{masson_phoneme_similarity_tts} show that artificial generation simulates the patterns of non-native speakers well in ASR systems. This would address one major shortcoming of any current ASR fairness study, which is unable to truly isolate individual speaker characteristics while selecting from a small population of speakers.






\section{Ethics statement}

Collecting recordings of minority groups, particularly children, requires care to avoid revealing their identities. Fair-speech avoids children altogether. Furthermore, our work is meant to increase fairness in ASR; however, by focusing on a small number of datasets, we potentially overlook SG's which face ASR discrimination, thereby reinforcing it. In this case we didn't consider patients with adverse health conditions, for example, a group which has been studied extensively in ASR fairness research and no less deserving of attention than those highlighted in our study \cite{parkinsons_asr}. By avoiding analyzing SG's for which benchmarks don't provide enough data, we are reinforcing the discrimination likely behind this unbalance in the first place.  

\FloatBarrier

\bibliographystyle{lrec2026-natbib}
\section{Bibliography}
\bibliography{these}

@inproceedings{aman2013speech,
  title = {Speech Recognition of Aged Voice in the {{AAL}} Context: {{Detection}} of Distress Sentences},
  shorttitle = {Speech Recognition of Aged Voice in the {{AAL}} Context},
  booktitle = {2013 7th {{Conference}} on {{Speech Technology}} and {{Human}} - {{Computer Dialogue}} ({{SpeD}})},
  author = {Aman, Frederic and Vacher, Michel and Rossato, Solange and Portet, Francois},
  year = 2013,
  month = oct,
  pages = {1--8},
  publisher = {IEEE},
  address = {Cluj-Napoca, Romania},
  doi = {10.1109/SpeD.2013.6682669},
  urldate = {2026-02-26},
  isbn = {978-1-4799-1065-6 978-1-4799-1063-2},
  langid = {english}
}

@inproceedings{amazon_kmeans,
  title = {Toward {{Fairness}} in {{Speech Recognition}}: {{Discovery}} and Mitigation of Performance Disparities},
  shorttitle = {Toward {{Fairness}} in {{Speech Recognition}}},
  booktitle = {Interspeech 2022},
  author = {Dheram, Pranav and Ramakrishnan, Murugesan and Raju, Anirudh and Chen, I.-Fan and King, Brian and Powell, Katherine and Saboowala, Melissa and Shetty, Karan and Stolcke, Andreas},
  year = 2022,
  month = sep,
  eprint = {2207.11345},
  primaryclass = {cs, eess},
  pages = {1268--1272},
  doi = {10.21437/Interspeech.2022-10816},
  urldate = {2024-04-15},
  archiveprefix = {arXiv}
}

@misc{ancogen_generation,
  title = {{{AnCoGen}}: {{Analysis}}, {{Control}} and {{Generation}} of {{Speech}} with a {{Masked Autoencoder}}},
  shorttitle = {{{AnCoGen}}},
  author = {Sadok, Samir and Leglaive, Simon and Girin, Laurent and Richard, Ga{\"e}l and {Alameda-Pineda}, Xavier},
  year = 2025,
  month = jan,
  number = {arXiv:2501.05332},
  eprint = {2501.05332},
  primaryclass = {cs},
  publisher = {arXiv},
  doi = {10.48550/arXiv.2501.05332},
  urldate = {2026-03-31},
  archiveprefix = {arXiv}
}

@inproceedings{arabic_asr,
  title = {The Effects of Speakers' Gender, Age, and Region on Overall Performance of {{Arabic}} Automatic Speech Recognition Systems Using the Phonetically Rich and Balanced {{Modern Standard Arabic}} Speech Corpus},
  booktitle = {Proceedings of the 2nd {{Workshop}} of {{Arabic Corpus Linguistics}}},
  author = {Abushariah, Mohammad and Sawalha, Majdi},
  year = 2013,
  month = jan,
  address = {Lancaster, UK}
}

@inproceedings{artie,
  title = {Artie {{Bias Corpus}}: {{An Open Dataset}} for {{Detecting Demographic Bias}} in {{Speech Applications}}},
  shorttitle = {Artie {{Bias Corpus}}},
  booktitle = {Proceedings of the {{Twelfth Language Resources}} and {{Evaluation Conference}}},
  author = {Meyer, Josh and Rauchenstein, Lindy and Eisenberg, Joshua D. and Howell, Nicholas},
  editor = {Calzolari, Nicoletta and B{\'e}chet, Fr{\'e}d{\'e}ric and Blache, Philippe and Choukri, Khalid and Cieri, Christopher and Declerck, Thierry and Goggi, Sara and Isahara, Hitoshi and Maegaard, Bente and Mariani, Joseph and Mazo, H{\'e}l{\`e}ne and Moreno, Asuncion and Odijk, Jan and Piperidis, Stelios},
  year = 2020,
  month = may,
  pages = {6462--6468},
  publisher = {European Language Resources Association},
  address = {Marseille, France},
  urldate = {2026-02-16},
  isbn = {979-10-95546-34-4},
  langid = {english}
}

@misc{asr_survey_modern,
  title = {Automatic {{Speech Recognition}} in the {{Modern Era}}: {{Architectures}}, {{Training}}, and {{Evaluation}}},
  shorttitle = {Automatic {{Speech Recognition}} in the {{Modern Era}}},
  author = {Nayeem, Md and Tabrej, Md Shamse and Deb, Kabbojit Jit and Goswami, Shaonti and Hakim, Md Azizul},
  year = 2025,
  month = oct,
  number = {arXiv:2510.12827},
  eprint = {2510.12827},
  primaryclass = {eess},
  publisher = {arXiv},
  doi = {10.48550/arXiv.2510.12827},
  urldate = {2026-02-16},
  archiveprefix = {arXiv}
}

@inproceedings{bias_environment,
  title = {Unveiling {{Biases}} While {{Embracing Sustainability}}: {{Assessing}} the {{Dual Challenges}} of {{Automatic Speech Recognition Systems}}},
  shorttitle = {Unveiling {{Biases}} While {{Embracing Sustainability}}},
  booktitle = {Interspeech 2024},
  author = {Kulkarni, Ajinkya and Kulkarni, Atharva and Couceiro, Miguel and Trancoso, Isabel},
  year = 2024,
  month = sep,
  eprint = {2503.00907},
  primaryclass = {cs},
  pages = {4628--4632},
  doi = {10.21437/Interspeech.2024-2494},
  urldate = {2026-02-15},
  archiveprefix = {arXiv}
}

@inproceedings{big_thorough_bias_survey,
  title = {Bias in {{Automated Speaker Recognition}}},
  booktitle = {2022 {{ACM Conference}} on {{Fairness Accountability}} and {{Transparency}}},
  author = {Hutiri, Wiebke Toussaint and Ding, Aaron},
  year = 2022,
  month = jun,
  eprint = {2201.09486},
  primaryclass = {cs},
  pages = {230--247},
  doi = {10.1145/3531146.3533089},
  urldate = {2026-01-04},
  archiveprefix = {arXiv}
}

@inproceedings{bing_youtube_dialect,
  title = {Effects of {{Talker Dialect}}, {{Gender}} \& {{Race}} on {{Accuracy}} of {{Bing Speech}} and {{YouTube Automatic Captions}}},
  booktitle = {Interspeech 2017},
  author = {Tatman, Rachael and Kasten, Conner},
  year = 2017,
  month = aug,
  pages = {934--938},
  publisher = {ISCA},
  doi = {10.21437/Interspeech.2017-1746},
  urldate = {2024-03-01},
  langid = {english}
}

@article{black_white_princeville,
  title = {Racial Disparities in Automated Speech Recognition},
  author = {Koenecke, Allison and Nam, Andrew and Lake, Emily and Nudell, Joe and Quartey, Minnie and Mengesha, Zion and Toups, Connor and Rickford, John R. and Jurafsky, Dan and Goel, Sharad},
  year = 2020,
  month = apr,
  journal = {Proceedings of the National Academy of Sciences},
  volume = {117},
  number = {14},
  pages = {7684--7689},
  publisher = {Proceedings of the National Academy of Sciences},
  doi = {10.1073/pnas.1915768117},
  urldate = {2024-02-29}
}

@misc{callhome_corpus,
  title = {{{CABank English CallHome Corpus}}},
  author = {{Linguistic Data Consortium}},
  year = 2013,
  publisher = {TalkBank},
  doi = {10.21415/T5KP54},
  urldate = {2026-02-13}
}

@article{children_development_age_2,
  title = {Speech {{Development Between}} 30 and 119 {{Months}} in {{Typical Children I}}: {{Intelligibility Growth Curves}} for {{Single-Word}} and {{Multiword Productions}}},
  shorttitle = {Speech {{Development Between}} 30 and 119 {{Months}} in {{Typical Children I}}},
  author = {Hustad, Katherine C. and Mahr, Tristan J. and Natzke, Phoebe and Rathouz, Paul J.},
  year = 2021,
  month = oct,
  journal = {Journal of Speech, Language, and Hearing Research},
  volume = {64},
  number = {10},
  pages = {3707--3719},
  publisher = {American Speech-Language-Hearing Association},
  doi = {10.1044/2021_JSLHR-21-00142},
  urldate = {2026-02-15}
}

@misc{common_voice,
  title = {Common {{Voice}}: {{A Massively-Multilingual Speech Corpus}}},
  shorttitle = {Common {{Voice}}},
  author = {Ardila, Rosana and Branson, Megan and Davis, Kelly and Henretty, Michael and Kohler, Michael and Meyer, Josh and Morais, Reuben and Saunders, Lindsay and Tyers, Francis M. and Weber, Gregor},
  year = 2020,
  month = mar,
  number = {arXiv:1912.06670},
  eprint = {1912.06670},
  primaryclass = {cs},
  publisher = {arXiv},
  doi = {10.48550/arXiv.1912.06670},
  urldate = {2024-06-12},
  archiveprefix = {arXiv}
}

@misc{dutch_quantifying,
  title = {Quantifying {{Bias}} in {{Automatic Speech Recognition}}},
  author = {Feng, Siyuan and Kudina, Olya and Halpern, Bence Mark and Scharenborg, Odette},
  year = 2021,
  month = apr,
  number = {arXiv:2103.15122},
  eprint = {2103.15122},
  primaryclass = {cs, eess},
  publisher = {arXiv},
  urldate = {2024-05-28},
  archiveprefix = {arXiv},
  langid = {english}
}

@inproceedings{dutch_wav2vec_whisper,
  title = {Uncovering {{Bias}} in {{ASR Systems}}: {{Evaluating Wav2vec2}} and {{Whisper}} for {{Dutch}} Speakers},
  shorttitle = {Uncovering {{Bias}} in {{ASR Systems}}},
  booktitle = {2023 {{International Conference}} on {{Speech Technology}} and {{Human-Computer Dialogue}} ({{SpeD}})},
  author = {Fuckner, Marcio and Horsman, Sophie and Wiggers, Pascal and Janssen, Iskaj},
  year = 2023,
  month = oct,
  pages = {146--151},
  publisher = {IEEE},
  address = {Bucharest, Romania},
  doi = {10.1109/SpeD59241.2023.10314895},
  urldate = {2026-02-15},
  copyright = {https://doi.org/10.15223/policy-029},
  isbn = {979-8-3503-2797-7},
  langid = {english}
}

@inproceedings{fairness_definitions,
  title = {Fairness Definitions Explained},
  booktitle = {Proceedings of the {{International Workshop}} on {{Software Fairness}}},
  author = {Verma, Sahil and Rubin, Julia},
  year = 2018,
  month = may,
  series = {{{FairWare}} '18},
  pages = {1--7},
  publisher = {Association for Computing Machinery},
  address = {New York, NY, USA},
  doi = {10.1145/3194770.3194776},
  urldate = {2024-04-10},
  isbn = {978-1-4503-5746-3}
}

@misc{fairness_survey_classic,
  title = {A Comparative Study of Fairness-Enhancing Interventions in Machine Learning},
  author = {Friedler, Sorelle A. and Scheidegger, Carlos and Venkatasubramanian, Suresh and Choudhary, Sonam and Hamilton, Evan P. and Roth, Derek},
  year = 2018,
  month = feb,
  number = {arXiv:1802.04422},
  eprint = {1802.04422},
  primaryclass = {stat},
  publisher = {arXiv},
  doi = {10.48550/arXiv.1802.04422},
  urldate = {2026-02-14},
  archiveprefix = {arXiv}
}

@inproceedings{fairness_traps,
  title = {Fairness and {{Abstraction}} in {{Sociotechnical Systems}}},
  booktitle = {Proceedings of the {{Conference}} on {{Fairness}}, {{Accountability}}, and {{Transparency}}},
  author = {Selbst, Andrew D. and Boyd, Danah and Friedler, Sorelle A. and Venkatasubramanian, Suresh and Vertesi, Janet},
  year = 2019,
  month = jan,
  series = {{{FAT}}* '19},
  pages = {59--68},
  publisher = {Association for Computing Machinery},
  address = {New York, NY, USA},
  doi = {10.1145/3287560.3287598},
  urldate = {2026-02-16},
  isbn = {978-1-4503-6125-5}
}

@misc{fairspeech_bad_methodology_asr,
  title = {{{FairASR}}: {{Fair Audio Contrastive Learning}} for {{Automatic Speech Recognition}}},
  shorttitle = {{{FairASR}}},
  author = {Kim, Jongsuk and Yu, Jaemyung and Kwon, Minchan and Kim, Junmo},
  year = 2025,
  month = jun,
  number = {arXiv:2506.10747},
  eprint = {2506.10747},
  primaryclass = {eess},
  publisher = {arXiv},
  doi = {10.48550/arXiv.2506.10747},
  urldate = {2025-12-04},
  archiveprefix = {arXiv}
}

@misc{fleurs,
  title = {{{FLEURS-R}}: {{A Restored Multilingual Speech Corpus}} for {{Generation Tasks}}},
  shorttitle = {{{FLEURS-R}}},
  author = {Ma, Min and Koizumi, Yuma and Karita, Shigeki and Zen, Heiga and Riesa, Jason and Ishikawa, Haruko and Bacchiani, Michiel},
  year = 2024,
  month = aug,
  number = {arXiv:2408.06227},
  eprint = {2408.06227},
  primaryclass = {cs},
  publisher = {arXiv},
  doi = {10.48550/arXiv.2408.06227},
  urldate = {2026-02-13},
  archiveprefix = {arXiv}
}

@misc{fouldsIntersectionalDefinitionFairness2019,
  title = {An {{Intersectional Definition}} of {{Fairness}}},
  author = {Foulds, James and Islam, Rashidul and Keya, Kamrun Naher and Pan, Shimei},
  year = 2019,
  month = sep,
  number = {arXiv:1807.08362},
  eprint = {1807.08362},
  primaryclass = {cs},
  publisher = {arXiv},
  doi = {10.48550/arXiv.1807.08362},
  urldate = {2026-02-14},
  archiveprefix = {arXiv}
}

@misc{french_data_collection_law,
  title = {Loi N{$^\circ$} 78-17 Du 6 Janvier 1978 Relative \`a l'informatique, Aux Fichiers et Aux Libert\'es},
  year = 1978,
  month = jan,
  number = {78-17 8.1}
}

@misc{gender_perf_gaps_multilingual_pebbles,
  title = {Twists, {{Humps}}, and {{Pebbles}}: {{Multilingual Speech Recognition Models Exhibit Gender Performance Gaps}}},
  shorttitle = {Twists, {{Humps}}, and {{Pebbles}}},
  author = {Attanasio, Giuseppe and Savoldi, Beatrice and Fucci, Dennis and Hovy, Dirk},
  year = 2024,
  month = oct,
  number = {arXiv:2402.17954},
  eprint = {2402.17954},
  primaryclass = {cs},
  publisher = {arXiv},
  doi = {10.48550/arXiv.2402.17954},
  urldate = {2025-12-15},
  archiveprefix = {arXiv}
}

@inproceedings{globe,
  title = {{{GLOBE}}: {{A High-quality English Corpus}} with {{Global Accents}} for {{Zero-shot Speaker Adaptive Text-to-Speech}}},
  shorttitle = {{{GLOBE}}},
  booktitle = {Interspeech 2024},
  author = {Wang, Wenbin and Song, Yang and Jha, Sanjay},
  year = 2024,
  month = sep,
  pages = {1365--1369},
  publisher = {ISCA},
  doi = {10.21437/Interspeech.2024-70},
  urldate = {2026-02-09},
  langid = {english}
}

@misc{hend_gender,
  title = {Exploring {{Gender Disparities}} in {{Automatic Speech Recognition Technology}}},
  author = {ElGhazaly, Hend and Mirheidari, Bahman and Moosavi, Nafise Sadat and Christensen, Heidi},
  year = 2025,
  month = feb,
  number = {arXiv:2502.18434},
  eprint = {2502.18434},
  primaryclass = {cs},
  publisher = {arXiv},
  doi = {10.48550/arXiv.2502.18434},
  urldate = {2025-03-11},
  archiveprefix = {arXiv}
}

@misc{ibm_single_axis,
  title = {{{AI Fairness}} 360: {{An Extensible Toolkit}} for {{Detecting}}, {{Understanding}}, and {{Mitigating Unwanted Algorithmic Bias}}},
  shorttitle = {{{AI Fairness}} 360},
  author = {Bellamy, Rachel K. E. and Dey, Kuntal and Hind, Michael and Hoffman, Samuel C. and Houde, Stephanie and Kannan, Kalapriya and Lohia, Pranay and Martino, Jacquelyn and Mehta, Sameep and Mojsilovic, Aleksandra and Nagar, Seema and Ramamurthy, Karthikeyan Natesan and Richards, John and Saha, Diptikalyan and Sattigeri, Prasanna and Singh, Moninder and Varshney, Kush R. and Zhang, Yunfeng},
  year = 2018,
  month = oct,
  number = {arXiv:1810.01943},
  eprint = {1810.01943},
  primaryclass = {cs},
  publisher = {arXiv},
  doi = {10.48550/arXiv.1810.01943},
  urldate = {2026-02-14},
  archiveprefix = {arXiv}
}

@article{intersectionality_first,
  title = {Demarginalizing the Intersection of Race and Sex: A Black Feminist Critique of Antidiscrimination Doctrine, Feminist Theory and Antiracist Policies},
  author = {Crenshaw, Kimberle},
  year = 1989,
  journal = {University of Chicago Legal Forum},
  volume = {1},
  pages = {139--167},
  langid = {english}
}

@inproceedings{intersectionality_ml_survey,
  title = {Towards {{Intersectionality}} in {{Machine Learning}}: {{Including More Identities}}, {{Handling Underrepresentation}}, and {{Performing Evaluation}}},
  shorttitle = {Towards {{Intersectionality}} in {{Machine Learning}}},
  booktitle = {2022 {{ACM Conference}} on {{Fairness Accountability}} and {{Transparency}}},
  author = {Wang, Angelina and Ramaswamy, Vikram V. and Russakovsky, Olga},
  year = 2022,
  month = jun,
  eprint = {2205.04610},
  primaryclass = {cs},
  pages = {336--349},
  doi = {10.1145/3531146.3533101},
  urldate = {2026-02-14},
  archiveprefix = {arXiv}
}

@inproceedings{laura_kmeans,
  title = {{Vers l'entra\^inement de mod\`eles de reconnaissance automatique de la parole auto-supervis\'es \'equitables sans \'etiquettes d\'emographiques}},
  booktitle = {{Actes des 32\`eme Conf\'erence sur le Traitement Automatique des Langues Naturelles (TALN), volume 1 : articles scientifiques originaux}},
  author = {{Alonzo-Canul}, Laura and Lecouteux, Benjamin and Portet, Fran{\c c}ois},
  editor = {Bechet, Fr{\'e}d{\'e}ric and Chifu, Adrian-Gabriel and {Pinel-sauvagnat}, Karen and Favre, Benoit and Maes, Eliot and Nurbakova, Diana},
  year = 2025,
  month = jun,
  pages = {780--790},
  publisher = {ATALA \textbackslash textbackslash\textbackslash textbackslash\& ARIA},
  address = {Marseille, France},
  urldate = {2026-04-20},
  langid = {fra}
}

@inproceedings{librispeech,
  title = {Librispeech: {{An ASR}} Corpus Based on Public Domain Audio Books},
  shorttitle = {Librispeech},
  booktitle = {2015 {{IEEE International Conference}} on {{Acoustics}}, {{Speech}} and {{Signal Processing}} ({{ICASSP}})},
  author = {Panayotov, Vassil and Chen, Guoguo and Povey, Daniel and Khudanpur, Sanjeev},
  year = 2015,
  month = apr,
  pages = {5206--5210},
  issn = {2379-190X},
  doi = {10.1109/ICASSP.2015.7178964},
  urldate = {2024-06-13}
}

@inproceedings{masson_phoneme_similarity_tts,
  title = {Investigating {{Phoneme Similarity}} with {{Artificially Accented Speech}}},
  booktitle = {Proceedings of the 20th {{SIGMORPHON}} Workshop on {{Computational Research}} in {{Phonetics}}, {{Phonology}}, and {{Morphology}}},
  author = {Masson, Margot and {Carson-berndsen}, Julie},
  editor = {Nicolai, Garrett and Chodroff, Eleanor and Mailhot, Frederic and {\c C}{\"o}ltekin, {\c C}a{\u g}r{\i}},
  year = 2023,
  month = jul,
  pages = {49--57},
  publisher = {Association for Computational Linguistics},
  address = {Toronto, Canada},
  doi = {10.18653/v1/2023.sigmorphon-1.6},
  urldate = {2026-01-04}
}

@misc{mengDontSpeakToo2022,
  title = {Don't Speak Too Fast: {{The}} Impact of Data Bias on Self-Supervised Speech Models},
  shorttitle = {Don't Speak Too Fast},
  author = {Meng, Yen and Chou, Yi-Hui and Liu, Andy T. and Lee, Hung-yi},
  year = 2022,
  month = apr,
  number = {arXiv:2110.07957},
  eprint = {2110.07957},
  primaryclass = {cs, eess},
  publisher = {arXiv},
  urldate = {2024-04-10},
  archiveprefix = {arXiv}
}

@misc{meta_fair,
  title = {Towards Measuring Fairness in Speech Recognition: {{Fair-Speech}} Dataset},
  shorttitle = {Towards Measuring Fairness in Speech Recognition},
  author = {Veliche, Irina-Elena and Huang, Zhuangqun and Kochaniyan, Vineeth Ayyat and Peng, Fuchun and Kalinli, Ozlem and Seltzer, Michael L.},
  year = 2024,
  month = aug,
  number = {arXiv:2408.12734},
  eprint = {2408.12734},
  primaryclass = {cs},
  publisher = {arXiv},
  doi = {10.48550/arXiv.2408.12734},
  urldate = {2025-02-14},
  archiveprefix = {arXiv}
}

@inproceedings{native_non-native_accent,
  title = {Leveraging Native Language Information for Improved Accented Speech Recognition},
  booktitle = {Interspeech 2018},
  author = {Ghorbani, Shahram and Hansen, John H. L.},
  year = 2018,
  month = sep,
  eprint = {1904.09038},
  primaryclass = {eess},
  pages = {2449--2453},
  doi = {10.21437/Interspeech.2018-1378},
  urldate = {2026-02-15},
  archiveprefix = {arXiv}
}

@misc{nltk,
  title = {{{NLTK}}: {{The Natural Language Toolkit}}},
  shorttitle = {{{NLTK}}},
  author = {Loper, Edward and Bird, Steven},
  year = 2002,
  month = may,
  number = {arXiv:cs/0205028},
  eprint = {cs/0205028},
  publisher = {arXiv},
  doi = {10.48550/arXiv.cs/0205028},
  urldate = {2026-04-07},
  archiveprefix = {arXiv}
}

@inproceedings{noise_means_bad_asr,
  title = {Analyzing the Performance of {{ASR}} Systems: {{The}} Effects of Noise, Distance to the Device, Age and Gender},
  shorttitle = {Analyzing the Performance of {{ASR}} Systems},
  booktitle = {Proceedings of the {{XX International Conference}} on {{Human Computer Interaction}}},
  author = {Rodrigues, Ana and Santos, Rita and Abreu, Jorge and Be{\c c}a, Pedro and Almeida, Pedro and Fernandes, S{\'i}lvia},
  year = 2019,
  month = jun,
  series = {Interacci\&\#xf3;n '19},
  pages = {1--8},
  publisher = {Association for Computing Machinery},
  address = {New York, NY, USA},
  doi = {10.1145/3335595.3335635},
  urldate = {2026-02-19},
  isbn = {978-1-4503-7176-6}
}

@article{old_voice_change,
  title = {How {{Does Our Voice Change}} as {{We Age}}? {{A Systematic Review}} and {{Meta-Analysis}} of {{Acoustic}} and {{Perceptual Voice Data From Healthy Adults Over}} 50 {{Years}} of {{Age}}},
  shorttitle = {How {{Does Our Voice Change}} as {{We Age}}?},
  author = {Rojas, Sandra and Kefalianos, Elaina and Vogel, Adam},
  year = 2020,
  month = feb,
  journal = {Journal of Speech, Language, and Hearing Research},
  volume = {63},
  number = {2},
  pages = {533--551},
  publisher = {American Speech-Language-Hearing Association},
  doi = {10.1044/2019_JSLHR-19-00099},
  urldate = {2026-02-16}
}

@inproceedings{parkinsons_asr,
  title = {Study of the {{Performance}} of {{Automatic Speech Recognition Systems}} in {{Speakers}} with {{Parkinson}}'s {{Disease}}},
  author = {{Moro-Vel{\'a}zquez}, Laureano and Cho, Jaejin and Watanabe, Shinji and {Hasegawa-Johnson}, Mark and Scharenborg, Odette and Kim, Heejin and Dehak, Najim},
  year = 2019,
  month = sep,
  pages = {3875--3879},
  doi = {10.21437/Interspeech.2019-2993}
}

@inproceedings{region_bias_scots,
  title = {Gender and {{Dialect Bias}} in {{YouTube}}'s {{Automatic Captions}}},
  booktitle = {Proceedings of the {{First ACL Workshop}} on {{Ethics}} in {{Natural Language Processing}}},
  author = {Tatman, Rachael},
  editor = {Hovy, Dirk and Spruit, Shannon and Mitchell, Margaret and Bender, Emily M. and Strube, Michael and Wallach, Hanna},
  year = 2017,
  month = apr,
  pages = {53--59},
  publisher = {Association for Computational Linguistics},
  address = {Valencia, Spain},
  doi = {10.18653/v1/W17-1606},
  urldate = {2024-02-29}
}

@misc{sensitive_attributes_automatic_stats,
  title = {A Statistical Approach to Detect Sensitive Features in a Group Fairness Setting},
  author = {Pelegrina, Guilherme Dean and Couceiro, Miguel and Duarte, Leonardo Tomazeli},
  year = 2023,
  month = may,
  number = {arXiv:2305.06994},
  eprint = {2305.06994},
  primaryclass = {cs},
  publisher = {arXiv},
  doi = {10.48550/arXiv.2305.06994},
  urldate = {2026-03-23},
  archiveprefix = {arXiv},
  langid = {english}
}

@article{smart_speakers_old_adults,
  title = {Smart {{Speakers}} for {{Health}} and {{Well-Being}} of {{Older Adults}}: {{A Mixed-Methods Review}}},
  shorttitle = {Smart {{Speakers}} for {{Health}} and {{Well-Being}} of {{Older Adults}}},
  author = {Dino, Michael Joseph and Leinbach, Carla and Dino, Gerald and Thiamwong, Ladda and Villafuerte, Chloe Margalaux and Shattell, Mona and Pimentel, Justin and Zamora, Maybelle Anne and Bautista, Anbel and Vitug, John Paul and Chepkorir, Joyline and Marave, Nerceilyn},
  year = 2025,
  month = jan,
  journal = {Healthcare},
  volume = {13},
  number = {21},
  pages = {2772},
  publisher = {Multidisciplinary Digital Publishing Institute},
  issn = {2227-9032},
  doi = {10.3390/healthcare13212772},
  urldate = {2026-02-13},
  copyright = {http://creativecommons.org/licenses/by/3.0/},
  langid = {english}
}

@article{snr_asr_mobile,
  title = {Noise-Robust Speech Recognition in Mobile Network Based on Convolution Neural Networks},
  author = {Bouchakour, Lallouani and Debyeche, Mohamed},
  year = 2022,
  month = mar,
  journal = {International Journal of Speech Technology},
  volume = {25},
  number = {1},
  pages = {269--277},
  issn = {1572-8110},
  doi = {10.1007/s10772-021-09950-9},
  urldate = {2026-04-07},
  langid = {english}
}

@inproceedings{solange_balanced,
  title = {Investigating the {{Impact}} of {{Gender Representation}} in {{ASR Training Data}}: A {{Case Study}} on {{Librispeech}}},
  shorttitle = {Investigating the {{Impact}} of {{Gender Representation}} in {{ASR Training Data}}},
  booktitle = {Proceedings of the 3rd {{Workshop}} on {{Gender Bias}} in {{Natural Language Processing}}},
  author = {Garnerin, Mahault and Rossato, Solange and Besacier, Laurent},
  editor = {{Costa-jussa}, Marta and Gonen, Hila and Hardmeier, Christian and Webster, Kellie},
  year = 2021,
  month = aug,
  pages = {86--92},
  publisher = {Association for Computational Linguistics},
  address = {Online},
  doi = {10.18653/v1/2021.gebnlp-1.10},
  urldate = {2024-05-28}
}

@inproceedings{sonos,
  title = {Sonos {{Voice Control Bias Assessment Dataset}}: {{A Methodology}} for {{Demographic Bias Assessment}} in {{Voice Assistants}}},
  shorttitle = {Sonos {{Voice Control Bias Assessment Dataset}}},
  booktitle = {Proceedings of the 2024 {{Joint International Conference}} on {{Computational Linguistics}}, {{Language Resources}} and {{Evaluation}} ({{LREC-COLING}} 2024)},
  author = {Sekkat, Chloe and Leroy, Fanny and Mdhaffar, Salima and Smith, Blake Perry and Est{\`e}ve, Yannick and Dureau, Joseph and Coucke, Alice},
  editor = {Calzolari, Nicoletta and Kan, Min-Yen and Hoste, Veronique and Lenci, Alessandro and Sakti, Sakriani and Xue, Nianwen},
  year = 2024,
  month = may,
  pages = {15056--15075},
  publisher = {{ELRA and ICCL}},
  address = {Torino, Italia},
  urldate = {2024-05-29}
}

@misc{speech_language_archive_analysis,
  title = {Global {{Performance Disparities Between English-Language Accents}} in {{Automatic Speech Recognition}}},
  author = {DiChristofano, Alex and Shuster, Henry and Chandra, Shefali and Patwari, Neal},
  year = 2023,
  month = feb,
  number = {arXiv:2208.01157},
  eprint = {2208.01157},
  primaryclass = {cs},
  publisher = {arXiv},
  doi = {10.48550/arXiv.2208.01157},
  urldate = {2026-03-23},
  archiveprefix = {arXiv}
}

@article{tech_prevalence_older_adults,
  title = {Digital Divide as a Determinant of Health in the {{U}}.{{S}}. Older Adults: Prevalence, Trends, and Risk Factors},
  shorttitle = {Digital Divide as a Determinant of Health in the {{U}}.{{S}}. Older Adults},
  author = {Yang, Rumei and Gao, Shiying and Jiang, Yun},
  year = 2024,
  month = dec,
  journal = {BMC Geriatrics},
  volume = {24},
  number = {1},
  pages = {1027},
  issn = {1471-2318},
  doi = {10.1186/s12877-024-05612-y},
  urldate = {2026-02-13},
  langid = {english}
}

@article{towards,
  title = {Towards Inclusive Automatic Speech Recognition},
  author = {Feng, Siyuan and Halpern, Bence Mark and Kudina, Olya and Scharenborg, Odette},
  year = 2024,
  month = mar,
  journal = {Computer Speech \& Language},
  volume = {84},
  pages = {101567},
  issn = {0885-2308},
  doi = {10.1016/j.csl.2023.101567},
  urldate = {2024-02-19}
}

@inproceedings{using_dat_vc_dutch,
  title = {Mitigating Bias against Non-Native Accents},
  booktitle = {Proc. {{Interspeech}} 2022},
  author = {Zhang, Yuanyuan and Zhang, Yixuan and Halpern, Bence and Patel, Tanvina and Scharenborg, Odette},
  year = 2022,
  pages = {3168--3172},
  doi = {10.21437/Interspeech.2022-836},
  urldate = {2024-02-26}
}

@misc{w2v2-xlsr,
  title = {{{XLS-R}}: {{Self-supervised Cross-lingual Speech Representation Learning}} at {{Scale}}},
  shorttitle = {{{XLS-R}}},
  author = {Babu, Arun and Wang, Changhan and Tjandra, Andros and Lakhotia, Kushal and Xu, Qiantong and Goyal, Naman and Singh, Kritika and von Platen, Patrick and Saraf, Yatharth and Pino, Juan and Baevski, Alexei and Conneau, Alexis and Auli, Michael},
  year = 2021,
  month = dec,
  number = {arXiv:2111.09296},
  eprint = {2111.09296},
  primaryclass = {cs},
  publisher = {arXiv},
  doi = {10.48550/arXiv.2111.09296},
  urldate = {2025-12-11},
  archiveprefix = {arXiv}
}

@misc{wav2vec2,
  title = {Wav2vec 2.0: {{A Framework}} for {{Self-Supervised Learning}} of {{Speech Representations}}},
  shorttitle = {Wav2vec 2.0},
  author = {Baevski, Alexei and Zhou, Henry and Mohamed, Abdelrahman and Auli, Michael},
  year = 2020,
  month = oct,
  number = {arXiv:2006.11477},
  eprint = {2006.11477},
  primaryclass = {cs, eess},
  publisher = {arXiv},
  urldate = {2024-01-23},
  archiveprefix = {arXiv}
}

@misc{whisper,
  title = {Robust {{Speech Recognition}} via {{Large-Scale Weak Supervision}}},
  author = {Radford, Alec and Kim, Jong Wook and Xu, Tao and Brockman, Greg and McLeavey, Christine and Sutskever, Ilya},
  year = 2022,
  month = dec,
  number = {arXiv:2212.04356},
  eprint = {2212.04356},
  primaryclass = {eess},
  publisher = {arXiv},
  doi = {10.48550/arXiv.2212.04356},
  urldate = {2025-02-18},
  archiveprefix = {arXiv}
}

@article{who_uses_smart_speakers,
  title = {Who Are the ({{Non-}}){{Adopters}} of {{Smart Speakers}}? {{A Cross-Sectional Survey Study}} of {{Dutch Families}}},
  shorttitle = {Who Are the ({{Non-}}){{Adopters}} of {{Smart Speakers}}?},
  author = {Wald, Rebecca and Piotrowski, Jessica Taylor and Van Oosten, Johanna M.F. and Araujo, Theo},
  year = 2024,
  month = mar,
  journal = {Tijdschrift voor Communicatiewetenschap},
  volume = {52},
  number = {1},
  pages = {4--28},
  issn = {1384-6930, 1875-7286},
  doi = {10.5117/TCW2023.X.001.WALD},
  urldate = {2025-04-15},
  langid = {english}
}


\clearpage
\appendix

\section{Regression with constituent DV's}

Another technique which has been used in the literature to estimate the effect of individual DV's on ASR performance is fitting a regression to predict ASR system's error on each speaker based on their constituent DV's \cite{sonos,region_bias_scots,speech_language_archive_analysis}. As in the previous case, it is imperative to fit this regression based on mean speaker performance rather than overall utterances in order to avoid biasing it towards individual speakers. The simplest form, regarding only univariate models and assuming categorical SG's, takes the following form:

\begin{align}
    \text{WER avg.}(D) = \sum\limits_{DV^i} \sum\limits_{SG \in DV^i}  \alpha_{SG} \cdot \mathbbm{1}_{(S = SG)} \label{eq:reg_uni}
\end{align}

\noindent where we expand $DV^i$ to include the everything-SG (to simulate a bias term $\alpha_{0}$). \citet{sonos} then goes on to define a multivariate model, which takes the intersection of SG's into account:

\begin{align}
    &\text{WER avg.}(D) = \nonumber \\
    &\big( \sum\limits_{DV^j}...\sum\limits_{DV^k} \sum\limits_{SG_j \in DV^j}...\sum\limits_{SG_k \in DV^k} \big) \nonumber \\
    &\alpha_{SG_j,...,SG_k} \cdot \mathbbm{1}_{(S = SG_j, ..., SG_k)} \label{eq:reg_multi}
\end{align}

\noindent which is the semantic equivalent of Section \ref{sec:isolated}. However, we consider relative SG-level WER to be a more intuitive measure; therefore, we focus on this for the remainder of the study.

\end{document}